# Design and Development of Robots' End Effector Test Rig

Josephine Selvarani Ruth D, Saniya Zeba, Vibha M R, Rokesh Laishram, Gauthama Anand

*Abstract*—A Test Rig for end-effectors of a robot is designed such that it achieves a prismatic motion in x-y-z axes for grasping an object. It is a structure, designed with a compact combination of sensors and actuators. Sensors are used for detecting presence, position and disturbance of target work piece or any object and actuators with motor driving system meant for controlling and moving the mechanism of the system. Hence, it improves the ergonomics and accuracy of an operation with enhanced repeatability.

## I. Introduction

Robots have been widely used in industries. With the advance technologies of humanoid robots, researchers believe it will help in replacing human work in difficult task and at the same time enhance work performance with less mistake. Conventional manipulators and end-effectors, which involve rigid components, such as linkages, gears, and motors, can exhibit precise positioning and improved mechanical performance using the control strategies and algorithms developed during the past decades. However, these types of rigid robots are unsafe and exhibit poor adaptability when they interact with humans or the surrounding environment.

The gripper testing facilities are designed and developed for robotic grippers or end effectors with suitable instrumentation such as Ultrasonic Sensor, Speed sensor, Force Sensor and Position Sensor. It is a motion platform structure which includes linear actuators, motors and sensors.

There are six main types of industrial robots: Cartesian, SCARA, cylindrical, delta, polar and vertically articulated. Each of these types offers a different joint configuration. However, there are several additional types of robot configuration of which Cartesian is more suitable as it has linear, prismatic motion which is a characteristic requirement for any pick and place operation.

Cartesian plane has movement in three axes (x, y and z) with lateral in x-axis, longitudinal in y axis and vertical in z-axis as shown in Fig. 1. The actuator moving in z-axis bears the gripper for grasping the object.

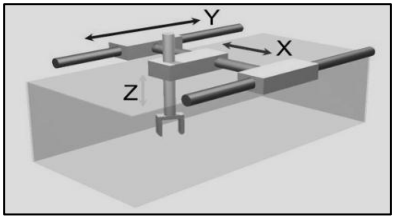

Fig. 1 Degrees of Freedom of a Cartesian Co-ordinate System

The different types of sensors used are as follows:

The Ultrasonic Sensor which is placed on the Test rig to detect any external disturbance or interference.

Force sensor is a two terminal device which is placed on the work piece and helps to determine the voltage to be applied to the gripper which in turn can be used to calculate the force for grasping the work piece.

The speed sensor is used to estimate the speed of the driving circuit of the actuator.

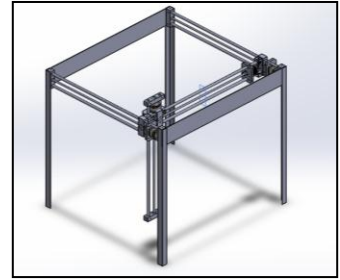

Fig. 2 Isometric View of the Test Rig

The model of the test rig was designed using modeling software called SolidWorks as shown in Fig. 2 and the load bearing members of the structure were analyzed using theories of mechanics of materials and Finite Element Method using ANSYS as a tool.

## II. Mathematical Analysis

### A. Equations

Consider a fixed-fixed stainless steel beam with uniform distributed load as shown in Fig. 3. It is a transverse load bearing member with torque maximum at the ends and minimum at the centre and the reaction forces with equal magnitude and acting in same direction at the two fixed ends, owing to the static equilibrium conditions as follows:

$$\sum F_x = 0 \quad \sum F_y = 0 \quad \sum F_z = 0$$

The analysis of fixed-fixed beam is as follows:

1. $R = \dfrac{wl}{2} \qquad M_{\max(end)} = \dfrac{wl^2}{12}$

*Research supported by DST INSPIRE & RBCCPS, Indian Institute of Science, Bangalore, Karnataka INDIA

Josephine S Ruth D, INSPIRE Faculty is with the Robert Bosch Centre for Cyber Physical Systems, Indian Institute of Science, Bengaluru -560 012.INDIA (djsruth@gmail.com, rhosephine@iisc.ac.in ).

Saniya Zeba and Vibha M R, Intern students and Rokesh Laishram, project staff are with the Robert Bosch Centre for Cyber Physical Systems, Indian Institute of Science, Bengaluru -560 012

2. $M_{centre} = \frac{wl^2}{24}$

Where,

R = Reaction force (N)

w = Load (N)

l = Length of the Stainless Steel rod (m)

M = Moment of Force (N-m)

All the units are in SI system.

B. *Calculations*

The calculations for the analysis of the load bearing member using the above equations is as shown below: From Eq. 1, $R = \frac{2 \times 9.81 \times 0.662}{2}$

$$R_A = R_B = 6.494N$$

From Eq. 2, $M_{max(end)} = \frac{2 \times 9.81 \times 0.662^2}{12}$

$M_{max(end)} = 0.7165 Nm$

From Eq. 3, $M_{centre} = \frac{2 \times 9.81 \times 0.662^2}{24}$

$M_{centre} = 0.3582 Nm$
From Eq. 1,
$w = \rho \times v \times g$
Where,

$\rho$ = Material Density of the component (kg/$m^3$)
v = Volume of the component ($m^3$)
g = Acceleration due to gravity (kg/$m^2$)

From the above equation, the total mass of an actuator acting as the uniformly distributed load on the load bearing member (Stainless Steel rod) is evaluated to be 2kgs.

The mass of individual components is as follows:
End Block = 0.44kg ; Mounting Plate = 0.16kg; Stepper Motor = 0.36kg; Shuttle = 0.21kg; Miscellaneous = 0.83kg
The maximum deflection of the beam at the center is as follows: $\Delta_{max(centre)} = \frac{wl^4}{384EI}$ Where,
$\Delta_{max(centre)}$ = Maximum Deflection at the Center (m)
I = Moment of Inertia of a beam with circular cross section ($m^4$)
By substituting the values in the above equation, we get,

$$\Delta_{max(centre)} = \frac{2 \times 9.81 \times 0.662^4}{384 \times 2 \times 10^{11} \times 2.15 \times 10^{-3}}$$

$$\Delta_{max(centre)} = 2.28 \times 10^{-11} m$$

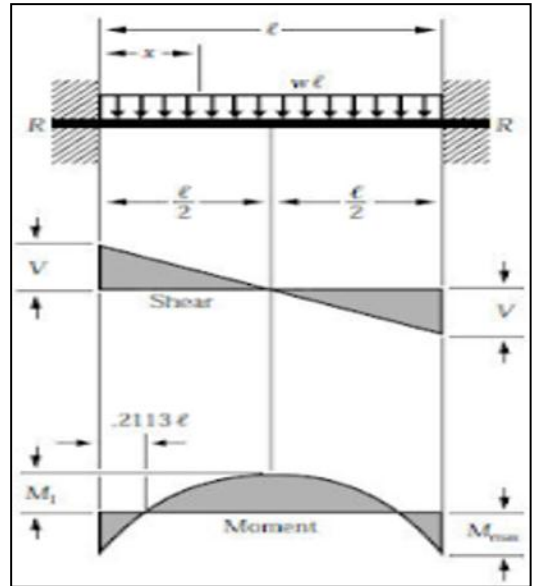

Fig. 3 Uniformly Distributed Loads on a Fixed-Fixed Beam

III. DESIGN AND CONSTRUCTION

There are various structural members in the Test Rig which are meant for bearing loads, supporting members, driving motors, fasteners, etc.

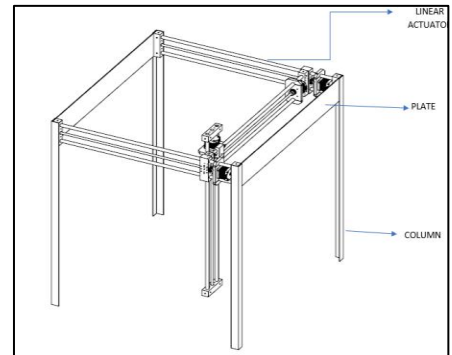

Fig. 4 Components of a Test Rig

As shown in Fig. 4, each linear actuator is mounted on the end blocks at both the ends and each end block on the Aluminum column.

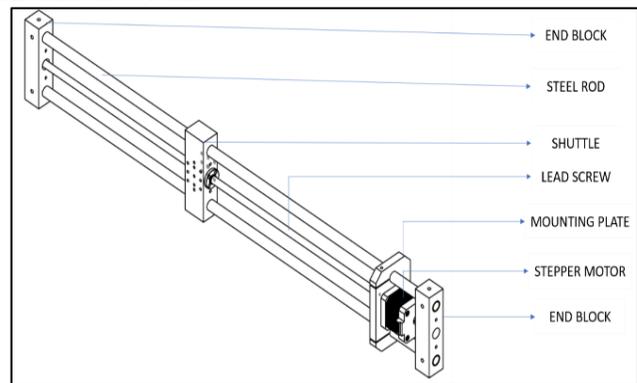

Fig. 5 Components of a linear actuator

As shown in Fig. 5, 700mm stationary, Stainless Steel rods with 12mm diameter and 2mm thickness placed on either side of the lead screw. The steel rods are considered to be the load bearing members, the analysis of which is proven later. A 600mm Trapezoidal four Start Lead Screw, material-304 Stainless Steel of 8mm Thread and 2mm Pitch with Brass Nut of 10 x 13 x 22 mm size is used. It has good wear resistance and strength, high accuracy, hard to rust and has good performances due to low friction coefficients and long usage spans. The reciprocating motion of the shuttle is achieved by the rotation of the threaded lead screw coupled to the stepper motor driving it.

The flexible type motor coupling can connect the driving shaft with the driven shaft while it is very efficient in eliminating any misalignment. Minimum backlash is another advantage of this. The outer diameter is 20mm, length 25mm

The ball bearing as shown in Fig. 6, has the bore diameter 8mm and 12mm outer diameter. It reduces rotational friction and supports radial and axial loads. The outer diameter is 20mm, length 25mm and bore dimensions 5×8mm.

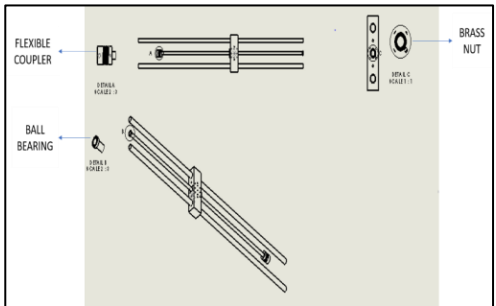
Fig. 6 Detailed Views of Coupler, Bearing and Nut

The End Block of dimensions 100×24×24mm is as shown in Fig. 8, is an Aluminum alloy. The function of this end block is to provide support to the actuator and reduce flexural and transverse vibrations of the load bearing members and also houses the ball bearing. The Shuttle of dimensions 100×24×32mm is an Aluminum alloy as shown in Fig. 9. The external threads on the Trapezoidal four start lead screw help the shuttle to achieve a smooth sliding motion between the end blocks. In addition to this, it acts as a surface of contact to the actuators to move in the other two axes. It is this motion of the shuttle that helps us to obtain a prismatic motion platform for the Test Rig. NEMA17 5.5kg-cm Stepper Motor is mounted on a Mounting Plate as shown in Fig. 7(b), of dimensions 100×60×10mm of Aluminum alloy, Fig. 10. The motor provides a torque at 1.5A current per phase. The motor's position can be commanded to move or hold at one position with the help of stepper motor drivers called RAMPS1.4 as shown in Fig. 7(a). RAMPS- RepRap Arduino Mega Pololu Shield consists of RAMPS1.4 shield, an Arduino Mega 2560 board and a maximum of five NEMA17 Stepper Drivers. It can control up to five stepper motors with as shown in Fig. 6.

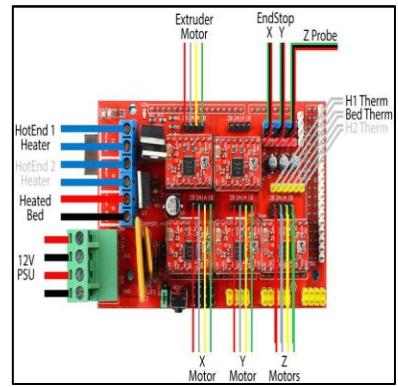
(a)

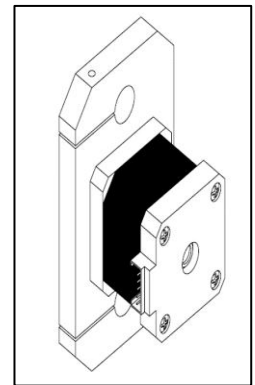
(b)

Fig. 7(a) RAMPS1.4   (b) NEMA17 Stepper Motor

## IV. COMPUTATIONAL ANALYSIS

The following are the results of analysis of fixed-fixed beam of a load bearing member which is the Stainless Steel rod in our case. Fig. 11 depicts the amplified deflection of the Stainless Steel rod of length 662mm (excluding the width of the end blocks) and outer diameter 12mm and inner diameter 10mm. The density of Stainless Steel is 7700kg/$m^3$ and Young's modulus is $2 \times 10^{11}$Pa. The Von Misses Stress, Displacement and Strain for a uniformly distributed load of 2kgs and 4kgs is as shown in the Fig. 12 and Fig. 13 respectively.

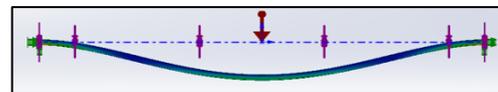
Fig. 11 Amplified Scale Deflection of Steel Rod

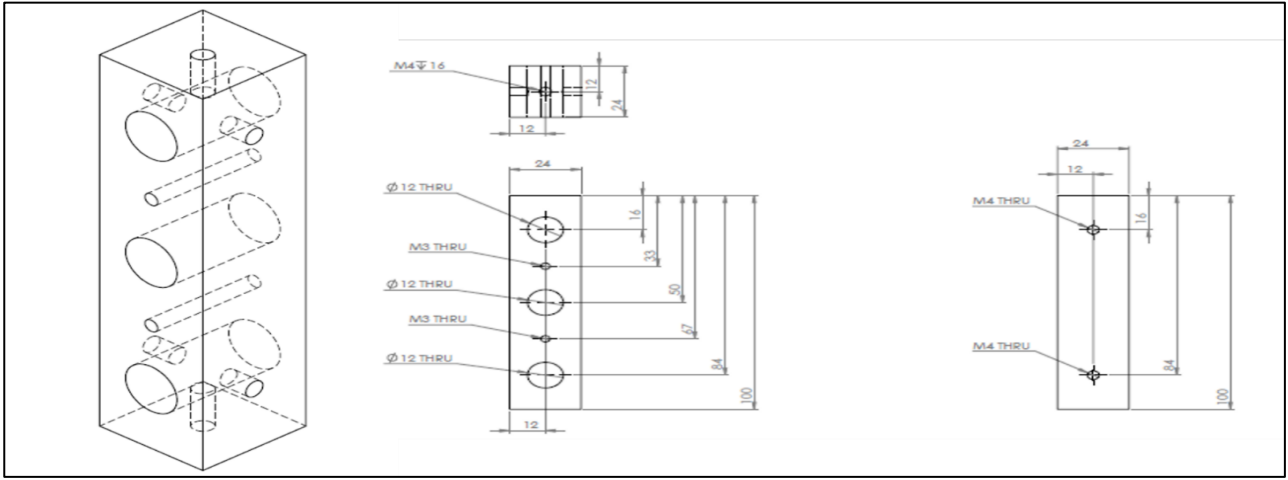

Fig. 8 Design of End Block

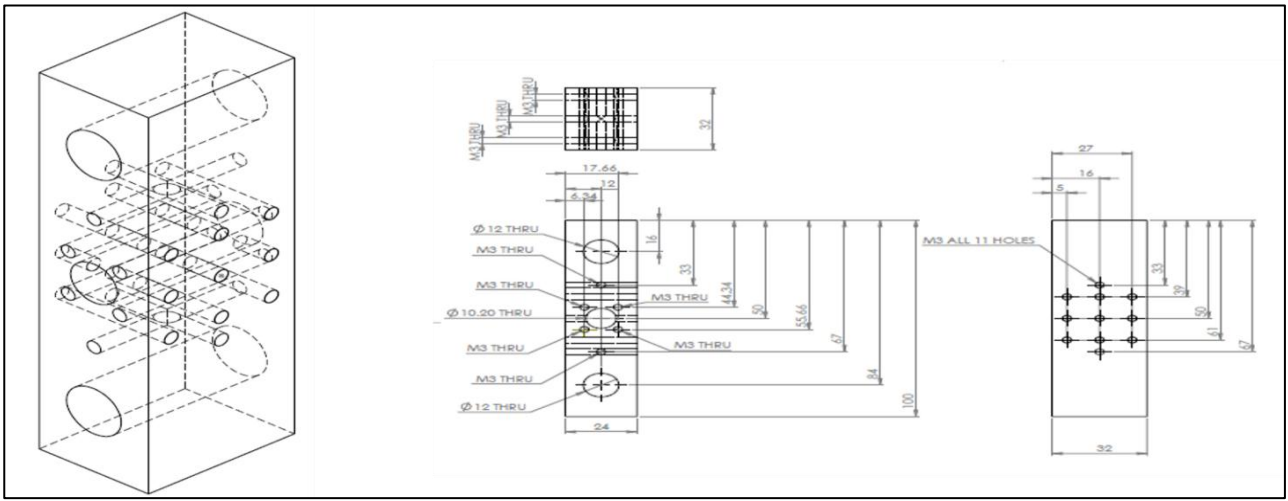

Fig. 9 Design of Shuttle

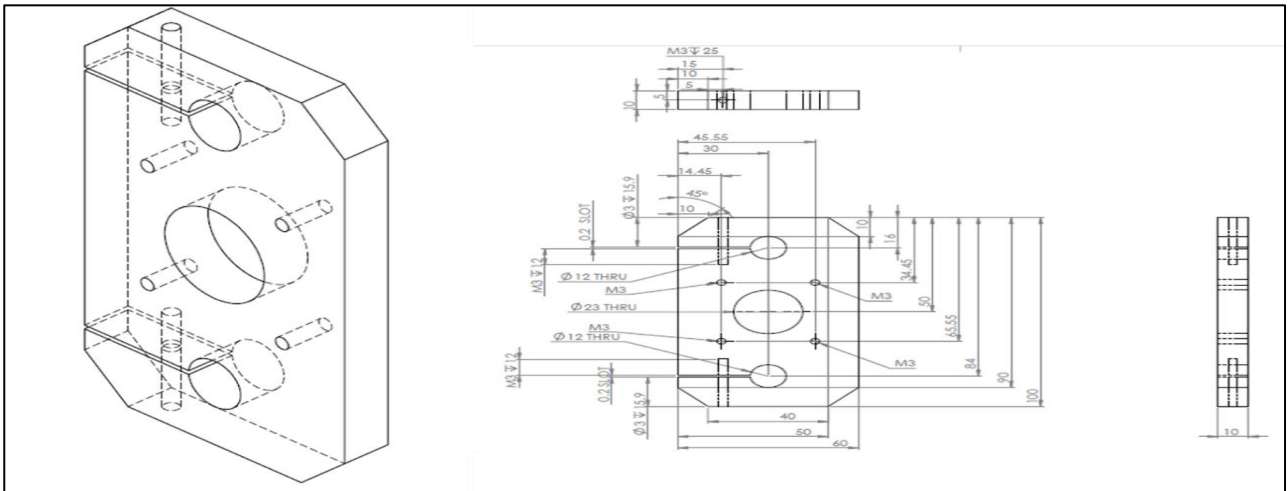

Fig. 10 Design of Mounting Plate

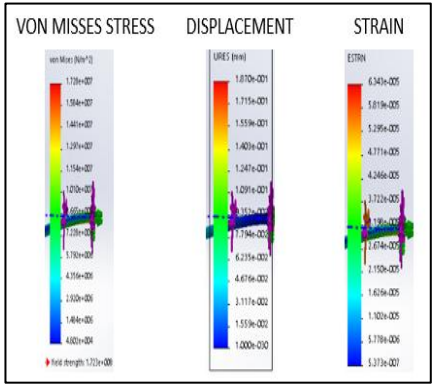

Fig. 12 Stress/Strain Values of 1×load

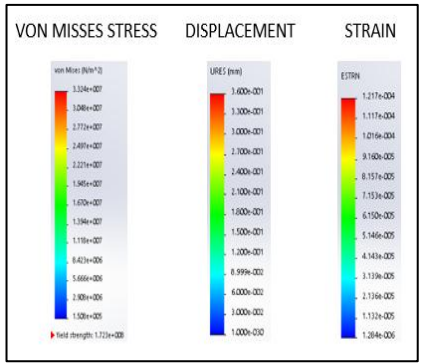

Fig. 13 Stress/Strain Values of 2×load

| Sl. No. | Mode | Frequency(Hz) |
| --- | --- | --- |
| 1 | 1 | 144.34 |
| 2 | 2 | 144.42 |
| 3 | 3 | 396.1 |
| 4 | 4 | 396.31 |
| 5 | 5 | 771.74 |

Table. 1 Natural Frequencies of Steel Rod

From the stress analysis, it is inferred that the maximum working stress is well within the yield strength. Hence, the load bearing member is safe from failure. The natural frequencies of the Stainless Steel rod of the dimensions as mentioned are obtained from Modal analysis using Finite Element Analysis with the help of ANSYS Workbench and are as shown in Fig. 14. The natural frequencies obtained for five modes are as shown in Table. 1.

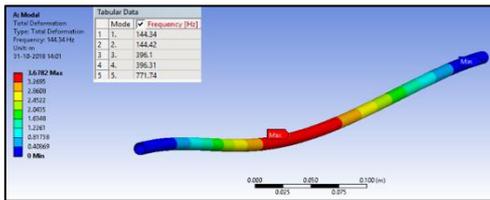

Fig. 14 Modal Analysis of Steel Rod

A harmonic analysis is used to determine the response of the structure under steady-state sinusoidal loading at a given frequency.

The following figures show the frequency response of stress and strain analysis done using harmonic analysis where the stress on the load bearing member is plotted against the operating frequencies. Three resonant frequencies were obtained between 0 and 300Hz. From Fig. 15, it is inferred that the member reaches its maximum fatigue stress of $2.55 \times 10^6$ N/$m^2$ at 100, 800 and 1800Hz. From the Fig. 16, it is inferred that the member reaches its maximum strain of $1.295 \times 10^{-5}$ at 100, 800 and 1800Hz.

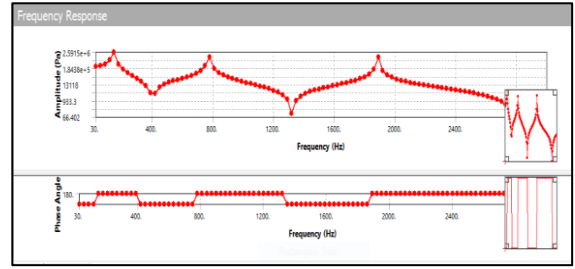

Fig. 15 Stress/Frequency Response of Steel Rod

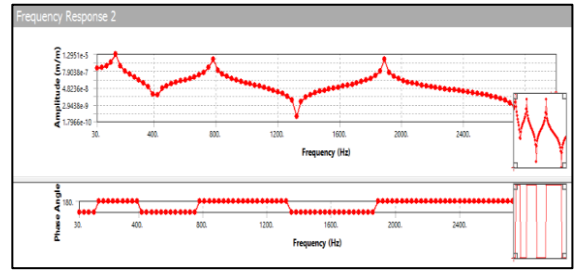

Fig. 16 Strain/Frequency Response of Steel Rod

## V. INSTRUMENTATION

### A. Generic Test Matrix

The main focus of constructing the test rig is to analyze the characteristics of the gripper designed. This is mainly to determine the gripper with the respective of their position with respect to eh target. In both static and dynamics conditions, the performance can be quantified with the three vital parameters as in the test matrix of the gripper.

Grasping force is the parameter which depends on weight of the object method of holding (physical construction or friction) coefficient of friction between fingers and object speed and acceleration during motion cycle. Operating bandwidth is the related to the optimal speed of operation when both are in dynamic condition; this is dependent in the actuation method of the griper that has been installed. Positioning efficiency is the factor in which the centre of gravity coincides overlaps when the gripper grasp and move with the target. A Fuzzy - PID controller is designed with three input members – target position, relative distance between the target and the end effectors (i.e.) depth the Z-axis actuator to slide to grasp the target, and speed of the test operation. The output of the fuzzy is the desired force that would be able to maintain a desired contact with test piece/target The fuzzy output provides the essential voltage, based on the rules developed from the previous empirical knowledge about the system, and is tuned for optimal performance. The controller uses the error and the rate of change of error as its inputs and meets the desired tuning parameters based on time-varying $e$ and $\dot{e}$. The schematic of the fuzzy – PID control is shown in Fig. 17.

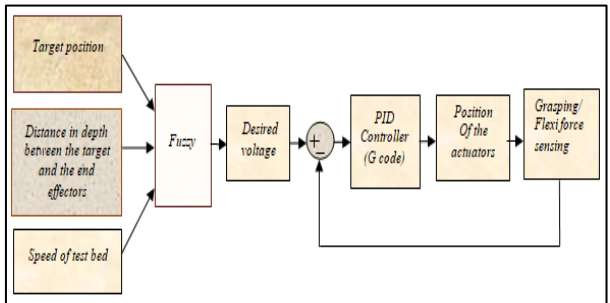

Fig. 17 Schematic of the controller for the gripper test rig system

*B. Sensors*

Sensors are a kind of transducers to convert energy from one form into another. They are used to detect and respond to electrical or optical signals. The different types of sensors used are Ultrasonic, Force and Speed sensors whose functions are elaborated in the next section.

The function of Ultrasonic sensor is similar to the principle of sonar that measure attributes of a particle by using the echoes from a radio or sound. It is also known as transceivers when the sensors both send and receive the signal. An ultrasonic testing produces high frequency sound waves and it measures the echo sent back by the sensor. Ultrasonic sensors calculate the time lapse between the sending wave and receiving echo to determine the distance to an object. A transmitter usually emits high frequency sound waves in the ultrasonic range (20 kHz–200 kHz), then change the electrical energy into sound. It is later change back into echo, and subsequently converted to electrical energy, which can be directly measured and displayed. This sensor will be placed on the Test rig to detect any external disturbance or interference.

This Force Sensitive Resistor which has an active sensing area of diameter 0.16″ (4 mm).will vary its resistance depending on how much pressure is being applied to the sensing area. Harder is the force, lower the resistance. When no pressure is being applied to the Force Sensitive Resistor, its resistance will be larger than 1MΩ and with full pressure applied the resistance will be 2.5kΩ. This sensor is placed on the work piece using which the voltage to be applied to the gripper can be determined, which in turn calculates the force for grasping the work piece.

Speed Measuring Sensor Groove Coupler Module for Arduino whose output interface can be directly connected to a micro-controller Input/output port if there is a block detection sensor, such as the speed of the motor encoder can detect. The modules can be connected to the relay, limit switch, and other functions. It is used in motor speed detection, pulse count, the position limit.

## VI. CONCLUSION

It has been shown that a full scale test rig can be designed to accurately impart dynamic environment for robot end effector To obtain its characteristics performance. The rig is based on the use of a standard servo test machine that loads the test panels. The design of the test rig is feasible for providing a motion platform for the gripper and also a conveyor belt system for the target. The application of this Test Rig is in pick and place operations of a target work piece placed on a conveyor belt as shown in Fig. 18.

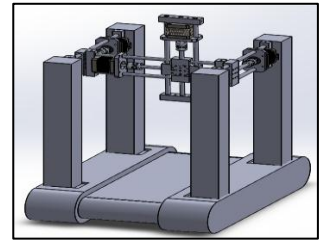

Fig. 18 Test Rig-Conveyor Belt System

ACKNOWLEDGMENT

The authors thank Varun V P, project staff at Robert Bosch Centre for Cyber Physical Systems, for his assistance in solid works that greatly improved the manuscript.
This work was financially supported in part by the GOVERNMENT OF INDIA MINISTRY OF SCIENCE & TECHNOLOGY, Department of Science & Technology under grant no. DST/INSPIRE/04/2017/000533. The work is hosted at Robert Bosch Centre for Cyber Physical Systems, Indian Institute of Science, Bengaluru -560 012, Karnataka INDIA.